\documentclass{article}

\usepackage{microtype}
\usepackage{graphicx}
\usepackage{subfigure}
\usepackage{booktabs} 

\usepackage{amsmath,amsfonts,bm}

\def\eqref#1{equation~\ref{#1}}

\def\1{\bm{1}}

\DeclareMathAlphabet{\mathsfit}{\encodingdefault}{\sfdefault}{m}{sl}
\SetMathAlphabet{\mathsfit}{bold}{\encodingdefault}{\sfdefault}{bx}{n}

\newcommand{\hatX}{\hat X}
\newcommand{\hatG}{\hat G}
\newcommand{\hatQ}{\hat Q}
\newcommand{\hatP}{\hat P}

\DeclareMathOperator*{\argmin}{arg\,min}

\usepackage{color,xcolor}
\usepackage{epsfig}
\usepackage{graphicx}

\usepackage{adjustbox}
\usepackage{array}
\usepackage{booktabs}
\usepackage{colortbl}
\usepackage{float,wrapfig}
\usepackage{hhline}
\usepackage{multirow}

\usepackage{amsmath,amsfonts,amsthm,amssymb}
\usepackage{bm}
\usepackage{nicefrac}
\usepackage{microtype}

\usepackage{changepage}
\usepackage{extramarks}
\usepackage{fancyhdr}
\usepackage{lastpage}
\usepackage{setspace}
\usepackage{soul}
\usepackage{xspace}

\usepackage{url}

\usepackage{algorithm, algorithmic}
\usepackage{enumerate}
\usepackage[draft]{todonotes} 

\usepackage{titlesec}

\newcolumntype{L}[1]{>{\raggedright\let\newline\\\arraybackslash\hspace{0pt}}m{#1}}
\newcolumntype{C}[1]{>{\centering\let\newline\\\arraybackslash\hspace{0pt}}m{#1}}
\newcolumntype{R}[1]{>{\raggedleft\let\newline\\\arraybackslash\hspace{0pt}}m{#1}}

\newcommand{\fig}[1]{Figure~\ref{#1}}

\newcommand{\ignore}[1]{}

\makeatletter
\DeclareRobustCommand\onedot{\futurelet\@let@token\@onedot}
\def\@onedot{\ifx\@let@token.\else.\null\fi\xspace}

\def\eg{e.g\onedot}

\def\etc{etc\onedot}

\makeatother

\definecolor{MyDarkBlue}{rgb}{0,0.08,1}
\definecolor{MyDarkGreen}{rgb}{0.02,0.6,0.02}
\definecolor{MyDarkRed}{rgb}{0.8,0.02,0.02}
\definecolor{MyDarkOrange}{rgb}{0.40,0.2,0.02}
\definecolor{MyPurple}{RGB}{111,0,255}
\definecolor{MyRed}{rgb}{1.0,0.0,0.0}
\definecolor{MyGold}{rgb}{0.75,0.6,0.12}
\definecolor{MyDarkgray}{rgb}{0.66, 0.66, 0.66}

\newcommand{\myparagraph}[1]{\vspace{-9pt}\paragraph{#1}}

\newcommand{\Model}{Visually Grounded Physics Learner\xspace}

\newcommand{\modelshort}{VGPL\xspace}

\usepackage{hyperref}

\usepackage[accepted]{icml2020}

\icmltitlerunning{Visual Grounding of Learned Physical Models}

\begin{document}

\twocolumn[
\icmltitle{Visual Grounding of Learned Physical Models}



\icmlsetsymbol{equal}{*}

\begin{icmlauthorlist}
\icmlauthor{Yunzhu Li}{mit_csail}
\icmlauthor{Toru Lin}{equal,mit_csail}
\icmlauthor{Kexin Yi}{equal,havard}
\icmlauthor{Daniel M.~Bear}{stanford_psy}
\icmlauthor{Daniel L.~K.~Yamins}{stanford_psy}\\
\icmlauthor{Jiajun Wu}{stanford_cs}
\icmlauthor{Joshua B.~Tenenbaum}{mit_bcs_cbmm_csail}
\icmlauthor{Antonio Torralba}{mit_bcs_cbmm_csail}
\end{icmlauthorlist}

\icmlaffiliation{mit_csail}{MIT CSAIL}
\icmlaffiliation{mit_bcs_cbmm_csail}{MIT BCS, CBMM, CSAIL}
\icmlaffiliation{havard}{Harvard University}
\icmlaffiliation{stanford_cs}{Department of Computer Science, Stanford University}
\icmlaffiliation{stanford_psy}{Wu Tsai Neurosciences Institute and Department of Psychology, Stanford University}

\icmlcorrespondingauthor{Yunzhu Li}{liyunzhu@mit.edu}

\icmlkeywords{Machine Learning, ICML}

\vskip 0.3in
]



\printAffiliationsAndNotice{\icmlEqualContribution} 

\begin{abstract}
Humans intuitively recognize objects' physical properties and predict their motion, even when the objects are engaged in complicated interactions. The abilities to perform physical reasoning and to adapt to new environments, while intrinsic to humans, remain challenging to state-of-the-art computational models. In this work, we present a neural model that simultaneously reasons about physics and makes future predictions based on visual and dynamics priors.
The visual prior predicts a particle-based representation of the system from visual observations. An inference module operates on those particles, predicting and refining estimates of particle locations, object states, and physical parameters, subject to the constraints imposed by the dynamics prior, which we refer to as \textit{visual grounding}. We demonstrate the effectiveness of our method in environments involving rigid objects, deformable materials, and fluids. Experiments show that our model can infer the physical properties within a few observations, which allows the model to quickly adapt to unseen scenarios and make accurate predictions into the future. 
\end{abstract}


\section{Introduction}

\begin{figure*}[t]
    \centering
    \includegraphics[width=\linewidth]{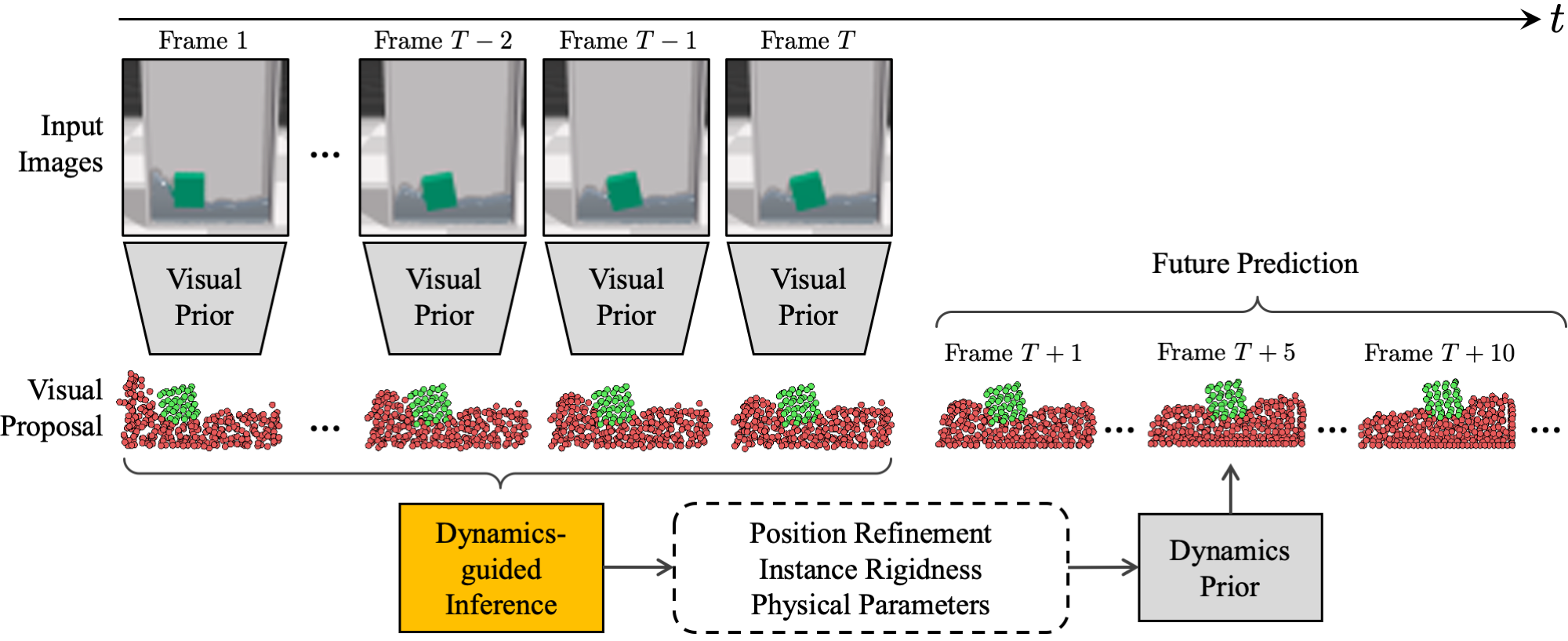}
    \vspace{-10pt}
    \caption{\textbf{Overview of \Model.} The model takes a sequence of image frames as input, reasons about the underlying physical properties, and makes future predictions. The input frames first go through the perception module ({\it visual prior}), which reconstructs the input scene in particle representation by giving a proposal on particle positions and instance groupings. The inference module then refines the proposal by updating the particle positions, estimating the rigidness of each instance, and predicting the physical parameters of the scene. The dynamics module ({\it dynamics prior}) takes in outputs from the inference module and predicts particle positions into the future. Please check our project page for video demonstrations.}
    \label{fig:model}
    \vspace{-5pt}
\end{figure*}

Understanding the physical properties of interacting objects has been a long-standing goal in computer vision, robotics, and artificial intelligence. As humans, by merely watching objects interact, we are able to distinguish between different object instances, reason about their physical properties, and make predictions on their future motion. More impressively, our ability to recognize, model, and predict the dynamics of physical systems
applies to not only rigid bodies, but also deformable objects such as elastic materials and fluids~\cite{bates2018modeling}. Given the example shown in Figure~\ref{fig:model}, humans can automatically
identify the separation between liquid (water) and solid (floating cube), estimate their properties such as gravity, density, and viscosity, and predict key features of their future motion through mental simulation~\cite{Battaglia2013Simulation,hamrick2016inferring}. 

For computational systems, physical reasoning on interacting deformable objects has been a highly challenging task, due to the diverse dynamical characteristics of different materials and their interactions. Take fluid as an example: fluids can deform, separate, merge, compress, and oscillate into arbitrary shapes, and some are hard to perceive due to their transparent nature. Prior works on system identification for robotics usually make strong assumptions on the structure of the underlying system~\citep{ljung2001system}, or require a lot of data to train a forward model~\citep{finn2017deep}, and therefore have a hard time modeling complex deformable objects like fluids and adapting to new scenarios. 

On the other hand, particle-based representations have recently gained attention in physical reasoning and control~\cite{mrowca2018flexible,hu2018chainqueen,li2019learning}.
Particles provide a dense and flexible representation that is well-suited for representing objects with diverse material and dynamical properties. Particles also facilitate relational inductive biases for more generalizable dynamics modeling. Recently, the DPI-Net~\cite{li2019learning}
has achieved strong results in modeling a variety of rigid and deformable object dynamics. The network is able to accurately predict the forward dynamics of the particles based on their pre-designated grouping and physical parameters (stiffness, viscosity, gravity, \etc). However, inference and visual grounding of these essential properties remain a challenging research problem.

In this work, we focus on the problem of physical reasoning about interacting deformable objects and propose a particle-based model that jointly refines the particle locations
and estimates their physical properties based on learned visual and dynamic priors. Our model, named \Model(\modelshort), first generates a coarse proposal of particle positions and grouping from raw visual observations (\textit{visual prior}). It then uses a learned dynamics model (\textit{dynamics prior}) to guide the inference of several essential system properties such as refinement of particle positions, object rigidness, and physical parameter estimation (\textit{dynamics-guided inference}). With those inferred quantities, our model can predict the future dynamics of the system.

We evaluate our model in environments involving interactions between rigid objects, elastic materials, and fluids. Experiments demonstrate that our model, within a few observation steps, is able to refine the particle positions proposed by the visual prior, accurately predict the rigidness the objects, and infer the physical parameters, which enables quick adaptation to new scenarios with unknown physical properties and making predictions into the future.

\section{Related Work}

Researchers have long been using neural networks to learn physical simulators~\citep{chen1990non,wan2001model}. Recently, people have demonstrated better generalization performance by using graph neural networks to capture the compositionality in dynamical systems~\citep{Battaglia2016Interaction,Chang2017compositional,sanchez2018graph,li2019propagation,li2020learning}. Other researchers have extended the model to particle systems, which showed impressive results in simulating objects made of different materials like rigid objects and fluids~\citep{mrowca2018flexible,li2019learning,Ummenhofer2020Lagrangian,sanchez2020learning}. These works make very few assumptions on the structure of the underlying systems, making them both general and flexible. Still, it remains as a question that how well they can handle raw visual inputs and adapt to environments of unknown physical properties.

\citet{Wu2015Galileo} introduced a method of inferring physical properties using MCMC, while others have tried differentiating through physics-based simulators to extract gradients~\citep{Todorov2012MuJoCo,drake,degrave2019differentiable,schenck2018spnets,hu2018chainqueen,de2018end,hu2019difftaichi,liang2019differentiable}, which showed strong results in solving inverse problems of various physical environments. However, their optimization process for dealing with the inverse problems is usually both time-consuming and prone to local optimum. Also, most of them directly operate on the state information of dynamical systems, lacking a way of handling raw visual inputs. This work aims to bridge the perception gap, enable physical reasoning from visual perception and perform dynamics-guided inference to directly predict the optimization results, which allows quick adaptation to environments with unknown physical properties.

People also have studied ways of reasoning about the physics and learning forward model directly from visual inputs~\citep{finn2017deep,babaeizadeh2018stochastic,hafner2018learning,ha2018world,Wu2017Learning}. However, these works either directly learn dynamics model over pixels or operate on a latent space, which limits their ability to reason about the physical properties explicitly and make accurate long time future predictions. Other researchers have shown better performance with intermediate representations like instance masks~\citep{Fragkiadaki2016Learning,watters2017visual,janner2018reasoning,yi2020clevrer}, object keypoints~\citep{minderer2019unsupervised}, or dense visual descriptors~\cite{DensePhysNet}. Instead, our model assumes a particle-based intermediate representation~\citep{macklin2013position}, allowing us to model interactions between objects of different materials, including rigid bodies, deformable objects, and fluids.

\section{Approach}

We present the \Model (\modelshort), a model that learns to infer the properties of a complex physical system guided by a learned dynamics model and grounded to visual inputs. \modelshort uses particles as the underlying state representation for physical modeling and inference. As shown in \fig{fig:model}, \modelshort first generates a coarse proposal of the particle states from input visual observations via a perception module ({\it visual prior}), including the positions and groupings of the particles. Our model then applies an inference module on these proposals, generating the refined positions of the particles, and estimating other physical properties such as object rigidity and physical parameters. Finally, we use a dynamics module ({\it dynamics prior}) to guide inference of these properties, which can predict future particle states from historical trajectories, conditioned on these properties. 
We describe details of \modelshort below.

\subsection{Problem Formulation}

Consider a system that contains $M$ objects and $N$ particles in its state representation. Given the visual observation $O = \{o^t\}_{t=1}^T$, our model first obtains a proposal of the particle position $\hatX'$ and the grouping information $\hatG$ for each particle, which is a probability distribution over the object instances, via a learned visual prior $f_V$. \modelshort also incorporates a learned dynamics prior $f_D$ that predicts future states based on the history of particle positions and physical properties of the system. These properties, including the rigidness of each object instance $\hatQ$ and the environmental physical parameters $\hatP$, are inferred by an inference module $f_I$. The inference module also generates a refinement $\Delta\hatX$ to the proposed particle locations. Our full model is summarized by the following equations:
\begin{align}
    & (\hatX', \hatG) = f_V(O), \\
    & (\hatP, \hatQ, \Delta\hatX) = f_I(\hatX', \hatG), \\ \label{eqn:inference}
    & \hatX = \hatX' + \Delta\hatX, \\
    & \hatX^{T+1} = f_D(\hatX, \hatG, \hatP, \hatQ). \label{eqn:dyn_pred}
\end{align}
The main objective of visual grounding is to infer the physical properties $(\hatP, \hatQ)$ and refine positions $\Delta\hatX$ from the visual proposals of the states, such that the dynamics model predicts the most accurate particle trajectories. Our inference module $f_I$ is tuned to minimize the following objective, constrained by fixed visual and dynamical priors $f_V$, $f_D$:
\begin{equation}
    (\hatP^*, \hatQ^*, \Delta\hatX^*) = \argmin\limits_{\hatP, \hatQ, \Delta\hatX} \|\hatX^{T+1} - X^{T+1}\|.
\end{equation}
In practice, the dynamics model iteratively predicts multiple steps into the future and this loss is computed over a finite time window.

\subsection{Visual Prior}

The visual prior proposes the particle state representation (position and grouping) from visual observations. The model architecture is built upon the point set generation network from \citet{Fan2017point}. Given a sequence of visual observation images $O = \{o^t\}_{t=1}^T$, the model first uses a convolutional encoder for extracting latent features and then applies two fully connected heads for predicting the position and grouping of the particles. The model outputs the normalized particle positions in each frame, as well as the probability distribution over all object instances that the particle might belong to, $(\hatX', \hatG) = f_V(O)$. In particular, the particles are in the $3$-dimensional space, $\hatX' = \{({x}_i^{t \prime}, {y}_i^{t \prime}, {z}_i^{t \prime})\}_{i=1, t=1}^{N, T}$ and $\hatG = \{G_i^t\}_{i=1, t=1}^{N, T}$ is a set of probability distributions over the object instances.

The visual prior is trained on the ground-truth particle states acquired from the physics engine. The full loss function is written as
\begin{equation}
    \mathcal{L_V} = \frac{1}{TN}\sum\limits_{t=1, i=1}^{T, N} \left[ \| {\hatX}_i^{t \prime} - X_i^t \|^2 + H(\hatG_i^t, G_i^t) \right],
\end{equation}
where $H$ stands for the cross entropy loss. 

In practice, in order to impose temporal consistency across different time steps, the network inputs a short sequence of images, and predicts the particle states over the same time window within a single pass. 

\subsection{Dynamics Prior}
\label{sec:approach_dyn_prior}

We adopt a particle-based dynamics model as the prior knowledge for guiding inference of the physical properties. At each time step, the positions of the particles $X$ define a point cloud that indicates the spatial span of the objects in the environment. The particles form groups $G$ to represent different object instances. Each particle has a binary rigidity label $Q$ that indicates whether the object it belongs to is a rigid body. Finally, the environment also has a set of real-valued physical parameters $P$, \eg, viscosity, gravity, stiffness, \etc.

\myparagraph{Physical state representation.}
To better model the time evolution of individual particle states and their interactions, we represent the physical state of the system with a graph $\langle V, E \rangle$. Each vertex $v_i \in V$ contains the position information of a single particle concatenated with the physical parameters, $v_i = (X_i, P)$. Each edge $(s, r) \in E$ contains a binary value $a_{sr} \in \{0, 1\}$ that indicates whether the sender $v_s$ and the receiver $v_r$ belong to the same object. Since the underlying interactions between the particles are local, at each time step the particles are connected to their neighbors within a specified distance threshold $d_e$.

\myparagraph{Spatial message passing.}
At each time step $t$, we use a graph neural network to perform the following updates on the graph representing the current physical state
\begin{align}
    g_{ij}^t = \phi_e(v_i^t, v_j^t, a_{ij})& \qquad (i, j) \in E \label{eqn:edge_update} \\  
    h_i^t = \phi_v(v_i^t, \sum_{j \in \mathcal{N}_i} g_{ji}^t)& \qquad i = 1, 2, \dots, N.  \label{eqn:vertex_update}
\end{align}
Here $\mathcal{N}_i$ is the set of all ``neighbors" of vertex $i$ with edges pointing to it. This process, which we refer to as \textit{spatial message passing}, also employed by many other physics modeling systems~\cite{Battaglia2016Interaction,sanchez2018graph}, generates a particle-centric encoding of the physical state $h_i^t$ at each vertex and time step. The same type of message passing on graph is also used in the inference module as we will discuss in Section~\ref{sec:approach_inference}. 

\myparagraph{Dynamics prediction.}
We use the dynamic particle interaction network (DPI-Net)~\cite{li2019learning} to perform dynamical update on the particle state based on the vertex embeddings obtained from spatial message passing. To incorporate temporal information, the network inputs multiple historical steps of the encoded physical state and predicts the particle positions one step ahead. The model handles rigid and non-rigid objects differently, therefore the update rules depend on the particles' grouping and rigidness:
\begin{equation}
    \hatX^{T+1}_i = \phi(\{h_i^t\}_{t=1}^T | G_i, Q_i).
\end{equation}
If a particle belongs to a rigid body, its motion can be decomposed into the translation of the body center plus the rotation of the particle with respect to the center. The update rule will apply a rigid body transformation (i.e. translation + rotation) to all particles belonging the same object to enforce the rigidity condition. For particles belonging to non-rigid objects, their position updates are independently computed per particle by a predictor network. Please refer to \citet{li2019learning} for further details.

\subsection{Dynamics-Guided Inference}
\label{sec:approach_inference}

\begin{figure*}[t]
\centering
\includegraphics[width=\linewidth]{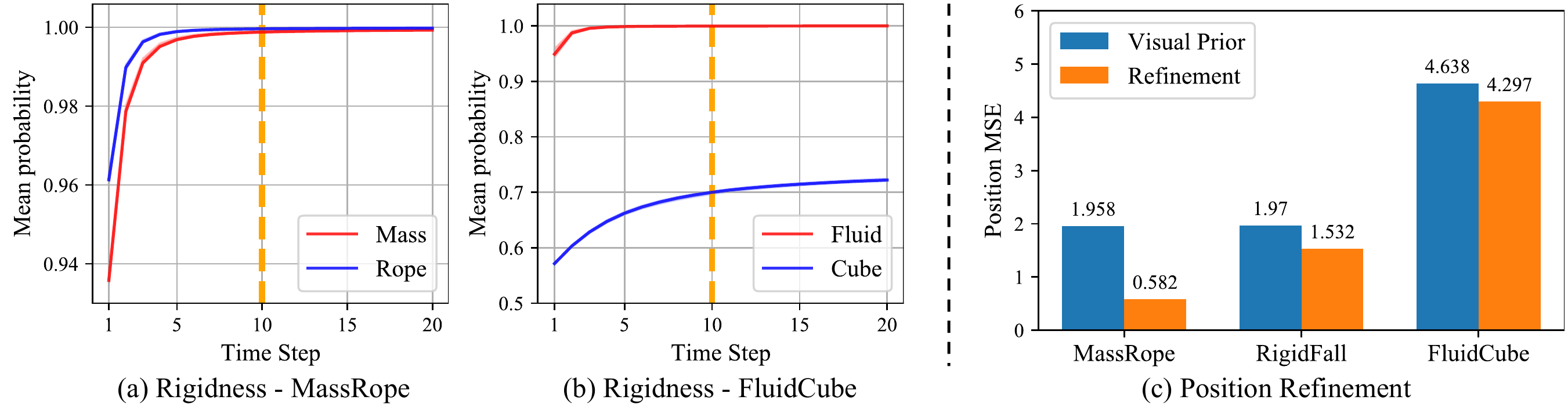}
\vspace{-10pt}
\caption{\textbf{Quantitative results on rigidness estimation and position refinement.} In (a) and (b), we show our model's performance on the rigidness estimation task in MassRope and FluidBox environments respectively. We use the mean probability of the ground truth rigidness label as the metric. The inference module was trained on inputs with only 10 time steps (the orange dashed line), but can extrapolate to both shorter and longer input sequence. Longer observation sequence leads to higher confidence, which is in line with our intuition. In (c), we show our model's performance on the position refinement task by comparing particle positions proposed by visual prior (in blue color) and after refinement by inference module (in orange color). We use the Mean Squared Error (MSE) between ground truth and predicted positions as the evaluation metric, scaled by $10^4$. In all environments, MSE decreases after refinement.}
\label{fig:rigidness_and_refinement}
\vspace{-10pt}
\end{figure*}
\begin{figure*}[t]
    \centering
    \includegraphics[width=\linewidth]{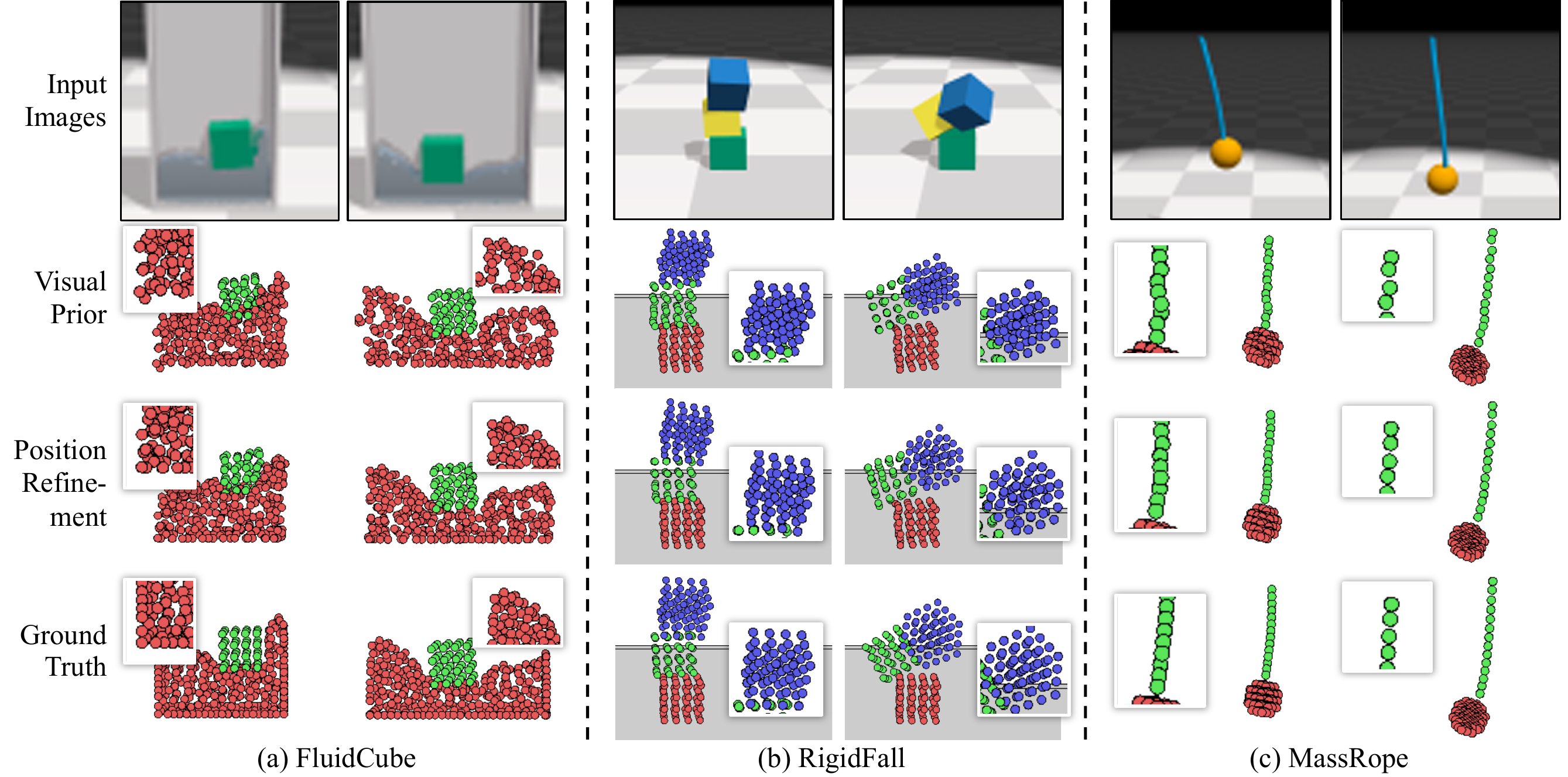}
    \vspace{-15pt}
    \caption{\small \textbf{Qualitative results on particle position refinement.} For each environment, we show side-by-side comparisons of two frames from the outputs of the visual prior, two frames from the outputs of position refinement, and two frames from the ground truth. For each output frame, we provide a zoom-in view to illustrate details of the particles. After refinement, (a) the fluids can better preserve the density constraint, (b) the rigid object is closer to the correct shape, and (c) the rope becomes less bumpy. The predicted particle positions after refinement all become closer to the ground truth.}
    \label{fig:vis_refine}
    \vspace{-15pt}
\end{figure*}

The key step of grounding a learned dynamics model to visual inputs is to infer the physical properties underlying the observed system, in our case, the rigidness of the objects $Q$, and the environmental physical parameters $P$. We apply an inference module $f_I$ to predict these properties from the observed particle proposals generated by the visual prior (equation \ref{eqn:inference}). The module also outputs refinement on the particles' positions $\Delta \hatX$. Since these properties are not directly accessible to the model, we use the learned dynamics prior to guide the inference of these properties. Details of the module are presented below.

\myparagraph{Spatiotemporal message passing.}

Given a sequence of length $T$ of the proposed particle positions and grouping $\hatX'$, $\hatG$, our inference module generates an embedding via \textit{spatiotemporal message passing}. Similar to the dynamics prior, the input physical state is represented by a graph $\langle V, E \rangle$, where at each vertex $v_i = (\hatX', \hatG)$ and the edges are connected between particles within a distance threshold $d_e$.

At each time step $t$, we first perform spatial message passing on the graph representation as described in Section~\ref{sec:approach_dyn_prior} to obtain the vertex embeddings $\{ h_i^t \}_{i=1}^N$ (equations \ref{eqn:edge_update}, \ref{eqn:vertex_update}). We then pass information on these embeddings along the temporal direction via a bi-directional recurrent network:
\begin{equation}
    u_i^t = \phi_{\tau}(\{h_i^{\tau}\}_{\tau=1}^T)^t \qquad t = 1, 2, \dots, T.
\end{equation}
In practice, we use the multi-layer perceptron (MLP) for $\phi_e$ and $\phi_{\tau}$, and the bi-directional Gated Recurrent Unit (GRU) \citep{chung2014empirical} for $\phi_{\tau}$. The weights of $\phi_{\tau}$ are shared across all vertices.

\myparagraph{Particle position refinement.}
We apply a refinement head $\phi_x$ on the spatiotemporal embedding $u_i^t$ to predict refinement $\Delta \hatX$ on each particle's position at each time step
\begin{equation}
    \Delta \hatX_i^t = \phi_x(u_i^t).
\end{equation}
In our model, $\phi_x$ is chosen to be a MLP whose weights are shared across all particles and time steps.

\myparagraph{Object rigidness estimation.}
To estimate the rigidness of each object in the system, a principled way is to start from embeddings that are associated with each object instances in the system. This is obtained by gathering the vertex embeddings from all particles belonging to the object and take the element-wise average to obtain an embedding vector of the object
\begin{equation}
    w_j^t = \sum_{i \in \mathcal{O}_j} u_i^t / |\mathcal{O}_j|, \qquad j = 1, 2, \dots, M,
\end{equation}
where $M$ is the number of objects in the system and $\mathcal{O}_j$ is the set of all particles belonging to the $j$th object, $\mathcal{O}_j := \{i | G_i = j\}$. This object embedding is then sent to a neural network to estimate the probability distribution on the rigidness 
\begin{equation}
    \hatQ_j^t = \phi_q(w_j^t).
\end{equation}
In our model, $\phi_q$ is a MLP with sigmoid output, shared across all object instances and time steps.

\myparagraph{Physical parameter estimation.}
Finally, we estimate the environmental physical parameters. Since the parameters are global, we use the full embeddings of all particles from all time steps and feed them together into a network $\phi_p$ to output a set of real numbers representing the estimated mean of the parameters, via $\hatP = \phi_p(\{u_i^t\}_{i=1,t=1}^{N,T})$.
In practice, we set $\phi_p$ to be an MLP with hyperbolic tangent output.

\myparagraph{Training.}
We use the pre-trained dynamics prior to guide the training of the inference module without access to the ground-truth of the inferred quantities. As shown in equation \ref{eqn:dyn_pred}, the dynamics model inputs the inferred quantities, including the refined position, grouping, rigidness, and physical parameters, and predicts the future positions of the particles. We take the $\mathcal{L}_1$ distance between the predicted positions and ground truth as the loss function.  The inference module is trained by stochastic gradient descent whose gradients are computed by back-propagating through the dynamics prior. Parameters of the dynamics prior stay frozen during training. 
\section{Experiments}
\label{sec:exp}

We study our framework under three environments that incorporate different types of objects and facilitate rich interactions. In this section, we show results and present ablation studies on various inference and prediction tasks.

\subsection{Environment}

We use NVIDIA FleX \citep{macklin2014flex}, a particle-based physics engine to generate all data for training and testing. The data includes visual observations and the corresponding particle states. For all three environments, we use 90\% of the data for training and 10\% for testing.

\myparagraph{RigidFall}
This environment simulates the motion and interaction of three rigid cubes. The cubes initially form a vertical stack with random noise added to their horizontal positions. The stack is released from above a rigid horizontal surface, and the cubes collide with one another as they fall under gravity. Each cube consists of 64 particles ($4 \times 4 \times 4$). The physical parameter of this environment is the gravitational acceleration, which is randomly sampled from $[-15.0, -5.0]$ for each simulation. The full dataset contains 5,000 simulations, each of which has 120 time steps.

\myparagraph{FluidCube}
In this environment, a rigid cube floats on top of a container of homogeneous fluid. The container can move horizontally to shake the fluid inside. During simulation, the container is initialized with a horizontal velocity of $0$ and assigned a random horizontal acceleration at each time step. The rigid block is consisted of 48 particles, and the fluid is consisted of 300 particles. The viscosity of the fluid is randomly chosen from the range $[1.0, 100.0]$. We generate 2000 samples, each of which has 300 time steps.

\myparagraph{MassRope}
In this environment, a rigid spherical mass is attached to an elastic rope whose upper end is pinned to an actuator that drives the rope's motion. We use positive $y$-direction as the upward direction, and the initial $xyz$-position of the actuator is $[0, 1, 0]$. The mass swings under a constant gravitational force and other internal forces such as rope tension. During simulation, the actuator at the upper end of the rope is assigned random accelerations along the horizontal plane (i.e. $x$- and $z$- directions), which also changes accelerations of the mass. The rigid mass is consisted of 81 particles, and the deformable rope is consisted of 14 particles. Rope stiffness is randomly chosen from range $[0.25, 1.20]$. We generate 3000 simulations, each of which includes 200 time steps.

\subsection{Implementation Details}

\begin{table}[t]
\vspace{-5pt}
\caption{{\bf Quantitative results on parameter estimation.} Below, we compare our model with DensePhysNet~\cite{DensePhysNet} and another model whose dynamics prior does not impose any constraints for rigid body motion. We measure the performance using the Mean Absolute Error (MAE) between each model’s prediction and the ground truth. The numbers show the percentage of the MAE error with respect to the maximum parameter range. Numbers in parentheses report the standard deviation.}
\label{tab:parameter_estimation}
\begin{center}\small

\tabcolsep=0.08cm
\begin{tabular}{l|ccc}
\toprule
Methods & MassRope & RigidFall & FluidCube \\
\midrule
DensePhysNet & 24.5\% (15.1) & 25.7\% (15.4) & 28.6\% (15.0)\\
Ours w/o Rigidness & 3.4\% (2.2) & 7.4\% (4.1) & 22.2\% (14.7) \\
\modelshort (ours) & {\bf 2.9\% (1.3)} & {\bf 3.7\% (2.7)}
 & {\bf 17.5\% (13.6)} \\
\bottomrule
\end{tabular}
\end{center}
\vspace{-15pt}
\end{table}

We present detailed model architecture and training paradigms below. All models are implemented in PyTorch~\citep{pytorch2019} and trained with the Adam optimizer~\citep{Kingma2015Adam}.

\myparagraph{Visual prior.}

Our visual prior network consists of a feature encoder and fully connected output heads. The feature encoder has 4 stacked convolutional blocks, with 32, 64, 128 and 256 channels. Each block includes one $3 \times 3$ convolutional layer with batch normalization and ReLU activation.
The prediction heads for particle position and grouping are both bi-layer MLPs with hidden size 2048.

The network is trained on the rendered visual observations of the system as well as the corresponding particle positions and grouping labels. We use a batch size of 50 and a learning rate of $10^{-4}$ to train the model for 2700 iterations on all environments. The particle positions are normalized. The sequence length of input and output data per forward pass is set to be 4. At inference time, given a sequence of input frames, we run the network on a sliding window over the sequence. In order to enforce temporal consistency, we move the window one step forward at a time and append the output at the last step of the moving window to the result sequence.

\begin{table*}[t]
\caption{{\bf Quantitative results on future prediction.} We show the Mean Squared Error (MSE) between the future predictions of the particle positions and the ground truth on all environments, scaled by $10^4$. We evaluate our model's performance by ablating on different aspects of the model: (1) without rigidness estimation, (2) without parameter estimation, and (3) without positions and groupings refinement. As shown in the table, with better and more thorough estimation of physical properties, we can predict the future positions more accurately, especially when making long-term predictions.}
\label{tab:future_pred}
\begin{center}\small

\begin{tabular}{l|cccc|cccc|ccc}
\toprule
\multirow{2}{*}{Methods} & \multicolumn{4}{c}{FluidCube} & \multicolumn{4}{c}{RigidFall} & \multicolumn{3}{c}{MassRope} \\
& $T + 1$ & $T + 5$ & $T + 10$ & $T + 20$ & $T + 1$ & $T + 5$ & $T + 10$ & $T + 20$ & $T + 1$ & $T + 5$ & $T + 10$ \\
\midrule
w/o Rigidness & {\bf 3.864} & 5.100 & 7.631 & 13.62 & 2.283 & 10.68 & 43.93 & 198.1 & 0.898 & 4.849 & 16.40 \\
w/o Refinement & 4.530 & 6.349 & 8.584 & 10.50 & 2.640 & 6.720 & 16.71 & 57.10 & 2.298 & 3.628 & 7.493 \\
w/o Param. Est. & 3.894 & 5.363 & 7.557 & 10.19 & {\bf 2.110} & 6.229 & 16.04 & 51.91 & 0.845 & 4.612 & 24.48 \\
\modelshort (ours) & 3.887 & {\bf 5.038} & {\bf 6.531} & {\bf 7.998} & 2.112 & {\bf 6.190} & {\bf 15.73} & {\bf 50.78} & {\bf 0.807} & {\bf 2.724} & {\bf 7.338} \\
\bottomrule
\end{tabular}
\end{center}
\vspace{-15pt}
\end{table*}

\myparagraph{Dynamics prior.}
We adopt the DPI-Net \citep{li2019learning} as the model for the dynamics prior. As explained in Section \ref{sec:approach_dyn_prior}, the network operates on a graph representation of the physical state at each time step and predicts the information at the vertices at the next step. The distance threshold for edge connection between two vertices is set to be $d_e = 0.08$. In practice, the vertex and edge information are first separately encoded by two 3-layer MLPs with hidden and output size 150 before sent to the propagator networks $\phi_e$, $\phi_v$ for spatial message passing. Both $\phi_e$ and $\phi_v$ include one fully connected layer with output size 150. The predictor heads for both rigid and non-rigid particles are 3-layer MLPs with hidden size 150 and ReLU activations.

Our dynamics prior is trained on the ground-truth particle trajectories for 10 epochs under each environment. We use a batch size of 4 and a learning rate of $10^{-5}$. The model observes 4 past time steps and predicts 1 time step into the future. The particle positions are normalized at input and denormalized at output.

\myparagraph{Inference module.}
We use the same network architecture in the inference module for feature encoding and spatial message passing as in the dynamics prior: 3-layer MLPs for vertex and edge embedding and single fully connected layer for message passing. All hidden and output layers have size 150. For temporal message passing, we use a bi-directional GRU with two hidden layers of size 150. The prediction heads are MLPs with 1 hidden layer of size 150 and output size determined by the dimensions of the prediction targets. The rigidness and parameter estimation heads have extra sigmoid and tanh output activations respectively.

Our inference module is trained for 2 epochs on each environment, using a batch size of 2 and a learning rate of $10^{-5}$. Length of the input and output sequences is set to be $T = 10$. The inference module is implemented by two separate networks with the same architecture: one for particle position refinement and rigidness prediction, and the other for physical parameter estimation. Each network takes in the proposals from the visual prior and predicts the desired variables, which, together with the ground-truth labels of the other network's output, are sent to the dynamics prior to predict particle positions. Loss is computed by comparing the predictions with the ground truth trajectories.

\subsection{Results}

\begin{figure*}[t]
    \centering
    \includegraphics[width=\linewidth]{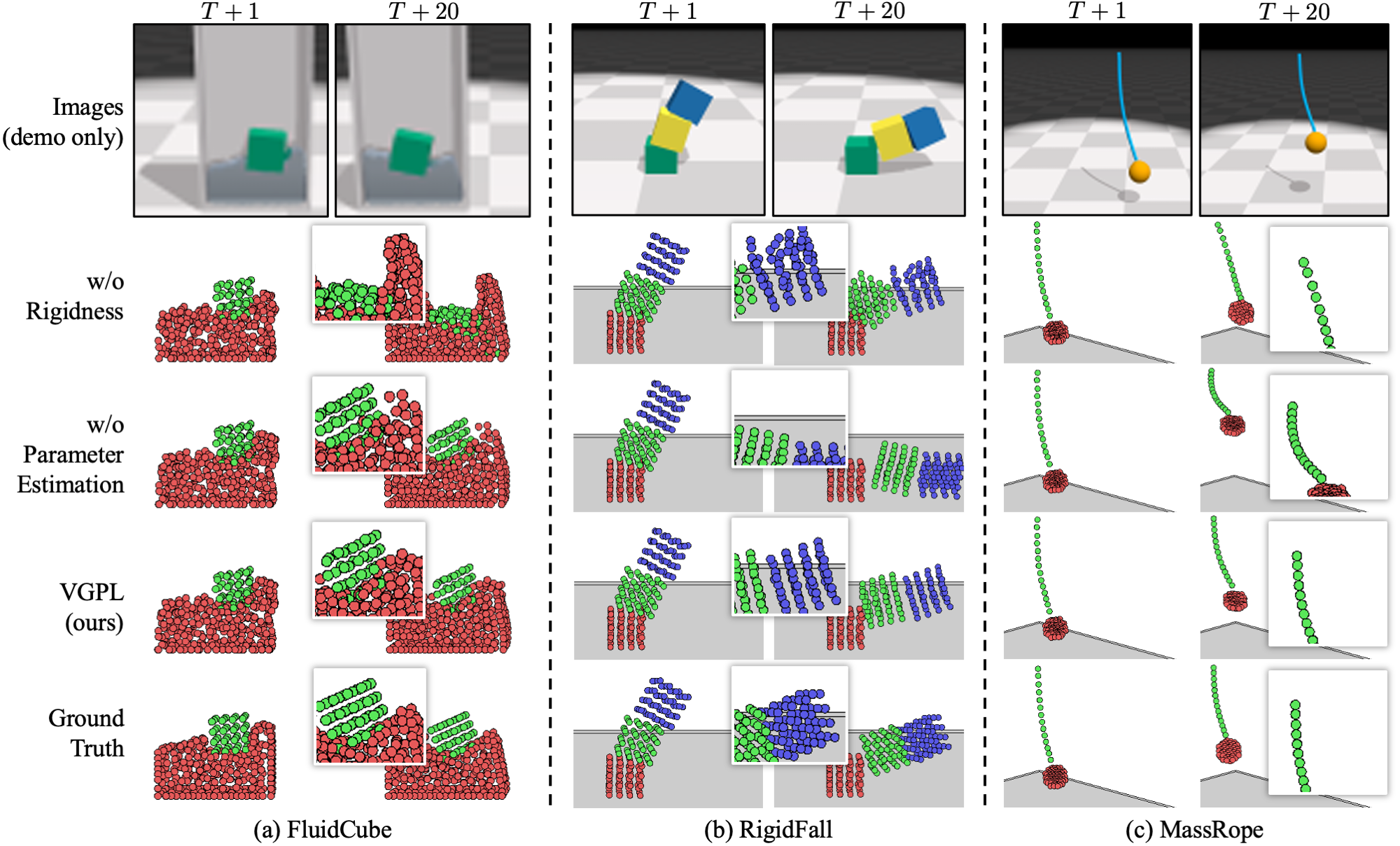}
    \vspace{-15pt}
    \caption{\textbf{Qualitative results on future predictions.} For each environment, we show results on predicted particle positions after $1$ and $20$ time steps. We compare the ground truth with the results of our model, together with versions without rigidness estimation or parameter estimation. For output frames after $20$ steps, we provide zoom-in views to show more details of the predicted particles. As shown in the figure, without proper estimation of the rigidness, (a) the rigid cube melts into the fluids and (b) the rigid cube scatters. Without an accurate estimate of the physical parameters, (b) the rigid boxes fall faster onto the ground, and (c) the rope contracts more than the ground truth. In all environments, our model performs the best, especially on the longer horizon.}
    \label{fig:vis_future_pred}
    \vspace{-10pt}
\end{figure*}

We evaluate the performance of \modelshort on the following tasks: accuracy of the inferred parameters including rigidness, position refinement and parameter estimation; and the prediction accuracy of future particle trajectories conditioned on these inferred properties. We also conduct ablation studies on these tasks to quantitatively evaluate the contributions from different model components.

\myparagraph{Rigidness estimation.}
In this task, we train the inference module on input sequences of length $T = 10$ and evaluate rigidness estimation on sequences of lengths $T =  1$ through $T = 20$. We focus on two environments for this task, MassRope and FluidCube, since RigidFall only contains rigid objects.

We use the mean probability weight on the correct rigidness label as the quantitative measure of inference accuracy. The model will choose the correct label at inference time if this probability is above 0.5. \fig{fig:rigidness_and_refinement}(a)(b) shows the relation between the inference accuracy and the input time steps for all object types existing in these environments. Our model achieves a good performance at input length 10 on all object types, especially for both objects in MassRope and fluid in FluidCube, with the mean probability close to 1.0. We also observe nice results under different input time steps. The mean probability further increases as the input length is increased beyond 10. This result shows temporal message passing in our inference module is generalizable to various input lengths. 

Our result also presents a notable gap between the mean probability of the cube versus the fluid in the FluidCube environment (\fig{fig:rigidness_and_refinement}(b)). This is due to the fact that the cube particles mostly move along the same direction as the fluid particles, and therefore are harder to recognize the rigidness. Intuitively, the rigidness of the cube becomes more obvious when it is moving against the water particles, not when it is ``riding the tide". 
As suggested by the result, a longer input sequence includes more opposite motion patterns between the cube and particle. It, therefore, leads to higher mean probability, which corresponds to higher confidence in the correct label of the rigidness.

\myparagraph{Position refinement.}
We evaluate position refinement via the deviation of the predicted positions from the ground truth trajectories. Figure \fig{fig:rigidness_and_refinement}(c) shows a quantitative comparison between the positions before and after the refinement. The result shows improvements in the mean squared error (MSE) on all environments, especially for MassRope where the MSE decreases by more than 3 fold.

We also show qualitative results in \fig{fig:vis_refine} to compare visualizations of the particles before and after refinement with the ground truth. As shown in the figure, in FluidCube, the fluid particle density becomes more uniform after the refinement, which is in agreement with the underlying assumption of the physics simulator that the incompressible fluid preserves density. In RigidFall, particle refinement is able to correct the deformation of the cube. This correction will largely affect the collision property of the cubes in dynamics modeling. In MassRope, the particles on the rope become less bumpy after the refinement.

\myparagraph{Physical parameter estimation.} 
DensePhysNet~\cite{DensePhysNet} has shown to be able to learn representations that carry rich physical information and can directly be used to decode physical object properties such as friction and mass. We compare with DensePhysNet by evaluating how well the models can estimate the physical parameters. We employ the same model and training procedure as used in DensePhysNet that iteratively takes the action and the current visual observation as input and tries to predict the optical flow, which is estimated using the algorithm developed by~\citet{liu2009beyond}. We then train a linear decoder that maps the resulting dense representation to the ground truth physical parameter.
On FluidCube, RigidFall, and MassRope, the parameters of interest are the fluid's viscosity, gravitational acceleration, and the stiffness of the rope, respectively. 

As shown in Table~\ref{tab:parameter_estimation}, where the numbers represent the percentage of the mean absolute error (MAE) with respect to the full range of these parameters, our model significantly outperforms DensePhysNet, showing the benefit of our formulation and the use of the visual and dynamics priors.
We also compare the result with another model whose dynamics prior treats all particles equally without imposing any constraints for rigid body motion. Our model shows higher accuracy on all environments, suggesting that a stronger dynamics model can provide better guidance to the inference module, and therefore lead to a more accurate estimation.

\myparagraph{Forward dynamics prediction.}
One important standard for judging the overall performance of visual grounding is the model's ability to accurately predict the future states of the system. To evaluate this, we send the inferred physical properties (i.e. position refinement, rigidness, and physical parameter) back to the dynamics prior and run forward pass on the network to iteratively predict the particles' future trajectories. We compute the mean squared error between the predicted particle positions and the ground truth after 1, 5, 10, 20 time steps as the quantitative benchmark to evaluate the performance over different time horizons (Table~\ref{tab:future_pred}). We also show qualitative results of predicted states in \fig{fig:vis_future_pred}.

To further study the impact of each inferred property on the overall performance, we perform ablation studies on each of the properties and compare with the full model. In the w/o rigidness model, the dynamics prior independently predicts the motion of each particle as non-rigid objects. This model predicts accurately within very short time horizon ($T + 1$) but fails after a few time steps as the rigid bodies melts into other shapes \fig{fig:vis_future_pred}(a)(b). In the w/o refinement model the dynamics prior inputs the coarse position proposals from the visual prior. This model shows poorer accuracy than the full model under all conditions due to the inaccurate inputs. The w/o parameter estimation model replaces the inferred parameter by a random number uniformly drawn from the parameter's range. Prediction of this model remains physically correct but deviates far from the ground truth at large time horizon \fig{fig:vis_future_pred}(b)(c). Overall, we show our full model achieves stronger performances than other baselines and demonstrate that all of the three inferred properties are essential to the task.
\section{Conclusion}

Humans have a strong ability to mentally simulate a variety of different substances, which helps us to distinguish between rigid and deformable objects and infer the material properties from visual observations. In this work, we propose a model, named \Model (\modelshort) that {\it grounds} the physical properties from vision, with the help of learned visual and dynamics priors. Our model employs a particle-based intermediate representation, which allows us to handle rigid bodies, deformable objects, and fluids. We have demonstrated in our experiments that our learned model can quickly adapt to new environments of unknown physical properties and make accurate predictions into the future.
\section*{Acknowledgements}

D.M.B. is supported by an Interdisciplinary Postdoctoral Fellowship from the Wu Tsai Neurosciences Institute and is a Biogen Fellow of the Life Sciences Research Foundation.

\bibliography{egbib,visual_grounding}
\bibliographystyle{icml2020}

\end{document}